\documentclass{article}
\usepackage[preprint]{neurips_2026}
\usepackage[utf8]{inputenc}
\usepackage[T1]{fontenc}
\usepackage{amsmath,amsfonts,amsthm}
\usepackage{booktabs}
\usepackage{graphicx}
\usepackage{microtype}
\usepackage{xcolor}
\usepackage{url}
\usepackage{hyperref}
\usepackage{placeins}
\hypersetup{colorlinks=true,linkcolor=blue,citecolor=blue,urlcolor=blue,
  pdftitle={Adaptive Reliability Reserves for Residential Battery Fleets},
  pdfauthor={Maruthi Vemula and Neeraj Praneeth Gajula}}
\newtheorem{proposition}{Proposition}
\author{%
  Maruthi Vemula\\
  University of North Carolina at Chapel Hill\\
  \texttt{vemula@unc.edu}
  \And
  Neeraj Praneeth Gajula\\
  New York University\\
  \texttt{Npg8461@nyu.edu}
}

\newcommand{\FixedPromise}{75.0\%}

\newcommand{\FixedUnserved}{204.0}
\newcommand{\ThresholdPromise}{86.5\%}
\newcommand{\ThresholdDelta}{+10.2}
\newcommand{\ThresholdLow}{+1.7}
\newcommand{\ThresholdHigh}{+21.7}
\newcommand{\ThresholdUnserved}{195.2}
\newcommand{\RiskPromise}{93.3\%}
\newcommand{\RiskDelta}{-25.4}

\newcommand{\LearnedPromise}{96.3\%}
\newcommand{\LearnedDelta}{-26.1}

\newcommand{\PromiseSeedLow}{31.1\%}
\newcommand{\PromiseSeedHigh}{96.3\%}
\newcommand{\PromiseSeedMedianDelta}{-99.5}
\newcommand{\PromiseCandidateCount}{10}
\newcommand{\PromiseBelowFixedCount}{3}
\newcommand{\PromiseNegativeMarginCount}{10}
\newcommand{\ThresholdGain}{10.2}
\newcommand{\ThresholdPromiseDelta}{11.4}
\newcommand{\ThresholdPromiseDeltaLow}{+0.0}
\newcommand{\ThresholdPromiseDeltaHigh}{+16.7}
\newcommand{\SelectedRun}{\texttt{ppo\_pc\_lam5\_s4}}
\newcommand{\TestFixedPromise}{95.8\%}
\newcommand{\TestThresholdDelta}{+5.1}
\newcommand{\TestThresholdPromise}{93.7\%}
\newcommand{\TestLearnedDelta}{-90.6}
\newcommand{\TestLearnedPromise}{98.6\%}

\newcommand{\PartialThresholdDelta}{+9.7}

\newcommand{\PartialLearnedDelta}{-66.1}

\newcommand{\DFWFixedPromise}{88.4\%}
\newcommand{\DFWThresholdPromise}{89.5\%}
\newcommand{\DFWThresholdDelta}{+12.3}
\newcommand{\DFWEpisodes}{95}
\newcommand{\SanAntonioFixedPromise}{97.7\%}
\newcommand{\SanAntonioThresholdPromise}{90.9\%}
\newcommand{\SanAntonioThresholdDelta}{+20.0}
\newcommand{\SanAntonioEpisodes}{44}
\newcommand{\UriFixedPromise}{90.9\%}

\newcommand{\UriRiskPromise}{94.4\%}

\newcommand{\UriLearnedPromise}{69.7\%}
\newcommand{\UriLearnedMargin}{-908.6}
\newcommand{\UriGuardedPromise}{98.0\%}
\newcommand{\UriGuardedMargin}{-785.4}
\newcommand{\MassEvents}{41}
\newcommand{\AlertCoverage}{85.4\%}
\newcommand{\AlertLeadCoverage}{19.5\%}
\newcommand{\SufficientTests}{74}
\newcommand{\FalseAlertWorstCost}{105.3}
\newcommand{\FalseAlertAfterDifference}{1.1}
\newcommand{\ACConvergence}{100.0\%}
\newcommand{\FixedETFade}{0.54\%}
\newcommand{\FixedBLASTFade}{4.40\%}
\newcommand{\HoldETFade}{0.39\%}
\newcommand{\HoldBLASTFade}{5.75\%}

\title{Adaptive Reliability Reserves\\for Residential Battery Fleets}

\begin{document}
\maketitle
\begin{abstract}
In a Houston 2024 replay, replacing a fixed 20\% battery reserve with a county-risk threshold raises observed five-hour backup coverage from \FixedPromise\ to \ThresholdPromise\ and adds \$\ThresholdGain\ in modeled annualized margin per home (paired 95\% interval: \$\ThresholdLow\ to \$\ThresholdHigh). The paired coverage-gain interval includes zero. This is a conditional simulation result, not measured company reliability. Residential batteries face this tradeoff because market discharge consumes energy also needed for household outages. We release HomeFleet-ERCOT, combining observed prices and county outages with identified simulated homes and causal controller inputs. We compare independent-home optimization with 21 proximal policy optimization (PPO) training runs; ten promise-cost candidates span \PromiseSeedLow--\PromiseSeedHigh\ backup coverage. Of these, \PromiseNegativeMarginCount\ have lower modeled margin than fixed 20\%, and \PromiseBelowFixedCount\ also have lower coverage. Chronological, geographic, fleet, and storm tests examine transfer. An alert charging override has a conditional energy-and-power sufficiency guarantee, but only \AlertLeadCoverage\ of \MassEvents\ county mass-outage episodes provide enough continuous warning for every battery preset. Synthetic tests include short, missed, and false warnings, load uplift, capacity loss, and dropout. A static AC audit retains existing feeder violations; aging comparisons expose model sensitivity. The released benchmark, controllers, and checkpoints provide a reproducible basis for comparing these tradeoffs. This work supports the design of residential battery programs that assess market value and household backup service together.
\end{abstract}

\section{Introduction}

A residential battery can reduce electricity costs while the grid is available and supply its household when the grid fails. The two uses compete for stored energy. Public operating descriptions illustrate the practical controls: Base Power describes retaining at least 20\% charge, while Tesla's Storm Watch charges toward maximum capacity during qualifying weather alerts~\citep{basecharge2026,teslastormwatch}. Those descriptions motivate our reference policies; our simulator does not reproduce either company's proprietary controller or customer outcomes.

We ask how much modeled market value a reserve can retain while protecting a short household backup commitment, and whether a learned decision rule continues to do so outside its training conditions. The reserve is the energy target retained while connected. The commitment is assessed separately: an outage episode succeeds if the battery serves its first five hours, or until earlier restoration, without unserved demand. A physically feasible trading action can still leave too little energy for this commitment.

\begin{figure}[t]
  \centering
  \includegraphics[width=\linewidth]{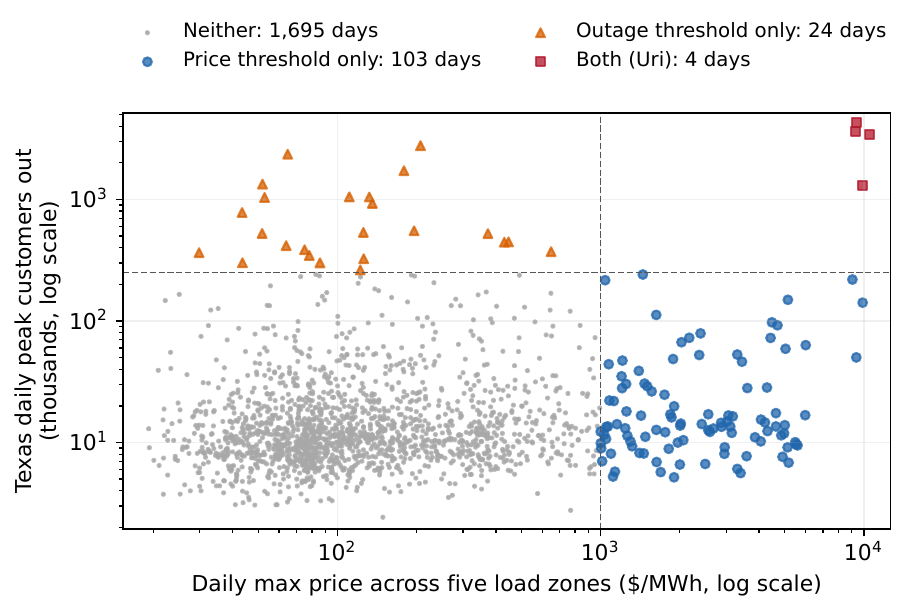}
  \caption{Historical price/outage context, 2021--2025. Each point is one local day. Dashed lines mark the two descriptive thresholds; both axes use logarithmic scales. The prices span Houston, North, South, Austin Energy, and CPS load zones. Statewide outage totals are not the household replay's reliability metric.}
  \label{fig:context}
\end{figure}

Historical timing motivates adaptation without establishing its safety. Across 1,826 days in 2021--2025, 107 days reached at least \$1,000/MWh in one of five ERCOT load zones, 28 reached at least 250,000 reported customers out statewide, and four met both thresholds; the four joint days occurred during Uri (Figure~\ref{fig:context}). This aggregate observation does not imply independence at a household, predict future storms, or bound the cost of charging during an alert. The rare joint event is precisely where an adaptive controller needs scrutiny.

Our contributions are three empirical and methodological pieces. First, we distinguish a per-household unserved-energy budget from the fixed-penalty surrogate actually implemented, and distinguish attainable battery constraints from outage-service guarantees. Second, HomeFleet-ERCOT supplies a versioned replay linking observed market/county records with identified simulated homes and causal controller inputs. Third, paired comparisons, complete seed reporting, named storms, synthetic warnings, and physical/accounting audits examine when a reserve policy's apparent advantage survives changes in conditions. The policy network and allocation mechanism are established components, rather than the claimed novelty.

The grid-facing setting is deliberately limited. The provider trades at exogenous zone prices, and its connected household dispatch changes distribution-level power exchange. We examine that exchange with static AC power flow. We do not model bulk-grid dynamics, endogenous market prices, or a network-constrained dispatch controller. This distinction makes the grid result and its limits assessable.

\section{Relation to prior work}

Residential storage's backup and economic uses have been studied under electricity tariffs~\citep{gorman2025backup}. Weather-informed home storage and microgrid control also precede this work~\citep{schmitz2020proactive,gutierrezrojas2021weather}. \citet{anunrojwong2026pooling} study coordination across residential batteries while preserving household backup commitments, using household telemetry and model predictive control. Here, household loads and interruptions are simulated from public sources, and the implemented optimization reference solves independent-home dynamic programs. We do not present it as a reproduction of their pooled controller. Learned storage reachability and terminal charge targets provide another related reserve mechanism~\citep{tapia2026reachability}.

Our learned controller uses proximal policy optimization (PPO)~\citep{schulman2017ppo} and a priority allocation/feasibility layer. Projection and correction of learned energy actions have extensive precedents~\citep{chen2021prof,ceusters2023safe,stanojev2023safe,karimimadahi2024control,markgraf2025projection}; realizable battery aggregation likewise requires individual power and energy accounting~\citep{elsaadany2026realizability}. These techniques motivate enforcing immediate operating constraints. They do not by themselves establish a household outage-service probability.

SustainGym emphasizes realistic energy-control evaluation under distribution shifts~\citep{yeh2023sustaingym}. Our benchmark focuses on the additional interaction between market dispatch, a local backup promise, and warning availability. Forecast assessment includes downstream economics and service outcomes, alongside classification metrics, because prediction accuracy alone need not determine decision value~\citep{lin2025accuracy}. Aging is assessed separately from dispatch, following the model-comparison motivation of \citet{humiston2026degradation}, with rainflow-based cycling~\citep{xu2018rainflow} and an SOC/temperature-dependent LFP model~\citep{blastlite}.

\section{Operating model and reliability objective}
\label{sec:model}

We study a provider that supplies household electricity and controls batteries installed at those households. The provider can lower its wholesale energy costs by discharging a battery when electricity is expensive. The same discharge reduces the energy available if that household subsequently loses grid power. Our decision variable is the amount of energy retained for that possible interruption. The setting is a simulation of a publicly described Texas retail and battery service, rather than a model fitted to an operator's private telemetry.

\subsection{Battery and market accounting}

Time advances in intervals of length $\Delta=0.25$ hours. Home $i$ has electricity demand $L_{i,t}$ and photovoltaic generation $V_{i,t}$, both in kW. Its battery contains $e_{i,t}$ kWh. Rated capacity is $\bar E_i$, and available capacity is $E_i=\bar E_i\,\mathrm{SOH}_i$, where state of health (SOH) is fixed within an episode. We use rated capacities of 25, 39.2, or 50 kWh, charge and discharge limits $\bar P_i^c$ and $\bar P_i^d$ of 11.4 or 20 kW, and charge and discharge efficiencies $\eta_c=\eta_d=\sqrt{0.9}$. With nonnegative charge and discharge powers $p^c_{i,t}$ and $p^d_{i,t}$,
\begin{equation}
 e_{i,t+1}=e_{i,t}+\Delta\left(\eta_c p^c_{i,t}-p^d_{i,t}/\eta_d\right).
\end{equation}
The implementation prevents simultaneous charging and discharging and enforces inverter limits and the physical energy range $[e_i^{\min},E_i]$, where $e_i^{\min}=0.05\bar E_i$. While connected, a home also has an operating reserve target $R_{i,t}$. If its current energy is below a newly raised target, the controller charges toward that target at the maximum feasible rate. This recovery can take several intervals; a target is not an instantaneous energy guarantee. Communication dropout prevents commanded grid-connected charging or discharging. Local islanded backup continues during communication dropout.

Let $p_{i,t}=p^d_{i,t}-p^c_{i,t}$, positive for discharge. Connected imports and exports are $m_{i,t}=(L_{i,t}-V_{i,t}-p_{i,t})^+$ and $x_{i,t}=(p_{i,t}-L_{i,t}+V_{i,t})^+$. Here $x^+=\max(x,0)$. Each fleet is evaluated within one ERCOT load zone $z$. Its settlement price $\lambda_{z,t}$ is converted from \$/MWh to \$/kWh. Following the accounting setting of \citet{anunrojwong2026pooling}, cash margin is
\begin{equation}
 \Pi_{i,t}=\Delta\left[p_{\rm ret}L^{\rm served}_{i,t}-(\lambda_{z,t}+c_{\rm TDSP})m_{i,t}+\lambda_{z,t}x_{i,t}-\beta V_{i,t}\right],
 \label{eq:margin}
\end{equation}
with retail price $p_{\rm ret}=0.09$, import delivery cost $c_{\rm TDSP}=0.05$, and solar credit $\beta=0.04$, all in \$/kWh. During islanding, imports and exports are zero and retail revenue uses served demand. Fixed subscription revenue is omitted from every comparison. These are gross modeled operating margins: they exclude hedging, capital costs, ancillary services, and degradation charges. A discharge that serves local demand avoids both the wholesale price and the delivery cost, whereas an export earns the wholesale price alone.

Prices are exogenous, so the fleet cannot change market prices. The dispatch model includes no transmission power flow, frequency dynamics, or market-clearing feedback. Section~\ref{sec:system} separately examines distribution-feeder voltages under realized dispatch using static AC power flow. That evaluation measures a physical consequence of synchronized dispatch; it does not make the energy controller network-constrained.

\subsection{Outage replay and the household promise}

For each county $c$, EAGLE-I provides the number of customers reported without electricity. We divide this count by the fixed modeled-customer denominator supplied with the dataset to obtain an outage fraction $f_{c,t}$. Within each contiguous county event above 0.2\%, each simulated home draws one $u_i\sim U(0,1)$ and is out whenever $u_i<f_{c(i),t}$. This construction reproduces the county fraction in expectation and couples each home's onset and restoration to the same event trace. It supplies simulated individual interruptions; the dataset does not contain observed interruptions for the sampled homes.

An islanded battery supplies positive demand net of available PV, subject to its discharge power limit and physical minimum. Any surplus PV can charge the battery. The operating reserve is released during islanding, and $U_{i,t}$ records unserved load in kWh. We use the entire household load as critical demand. For scoring, outage runs separated by at most one hour are merged into one episode. Training and evaluation use the same definition.

The \emph{promise rate} is the fraction of simulated outage episodes whose first five hours, or the time until earlier restoration, have zero unserved energy. Higher is better. This metric asks whether a short backup commitment is fulfilled, rather than whether a finite battery can serve a hurricane-length interruption in full. We also report total unserved energy and complete-episode ride-through. A split with no simulated outages has an undefined promise rate, reported as unavailable rather than as perfect reliability.

\subsection{Risk budgets and their practical relaxation}

A household reliability budget can be written as
\begin{equation}
 \mathbb E_\pi\!\left[\sum_{t\in Y}U_{i,t}\right]\le B_i \quad\text{for every home }i,
 \label{eq:budget}
\end{equation}
where $B_i$ is an allowable annual amount of unserved backup energy. Its Lagrange multiplier, $\lambda_{r,i}$, has units of \$/kWh and therefore acts as an implied value of lost load. Residential interruption-cost estimates provide context for its scale~\citep{ice2025,brown2024reliability}.

Our experiments use a shared, fixed multiplier $\lambda_r$ and sweep its value. They do not solve for household-specific multipliers or enforce every constraint in Equation~\eqref{eq:budget}. The budget motivates the reserve decision and the reported tradeoff; it is not a certificate of individual reliability. We distinguish the full unserved-energy penalty from a \emph{promise-window penalty}, which charges only for unserved energy in the first five hours of a scored episode. The latter changes the learning objective to match the short customer-facing promise. The model-based risk penalty below is a forecast shortfall surrogate, not an exact equivalent of either realized penalty.

\section{Public replay data and controller information}
\label{sec:data}

HomeFleet-ERCOT combines observed market prices and county outage records with simulated housing profiles. Each metropolitan pool contains 300 ResStock single-family homes. Houston uses Harris County; Dallas--Fort Worth uses Dallas, Tarrant, Collin, and Denton Counties; San Antonio uses Bexar County. ResStock AMY2018 load shapes are mapped to other years by local month, day, and clock time, with leap-day handling. They do not reproduce unusually high heating demand during Uri. Each home independently receives synthetic PV with probability 0.3, with a 5--9 kW DC rating and generation derived from irradiance and temperature. No battery-company household telemetry is included.

Real-time and day-ahead prices come from ERCOT, county outages from EAGLE-I~\citep{brelsford2024eaglei,eaglei_v4}, and building profiles from ResStock~\citep{resstock2024}. Weather is historical Open-Meteo model/reanalysis data used as an observation proxy~\citep{openmeteo_archive}, and warnings come from the Iowa Environmental Mesonet's NWS VTEC archive~\citep{iem_vtec}. Hourly weather is ingested as UTC instants before conversion to local time, preserving both occurrences of the repeated fall hour. Missing county observations are filled with zero for replay, an assumption that can understate outages when reporting coverage is incomplete. The data release includes sampled building identifiers, field definitions, input hashes, frozen bundles, and the code used to construct them.

\paragraph{Prices and loads.}
Controllers use lagged real-time prices, assuming the preceding quarter-hour's price is available at decision time; additional publication or telemetry delay is not modeled. Published day-ahead prices are available for the current delivery day and, from 14:00 local time, the following day. Unpublished future prices use per-slot medians from the preceding 28 days. Load and PV forecasts use preceding observations at the same local-time slots. Neither the initialization nor an update uses the evaluation episode's future realizations. The oracle reference is the explicitly identified exception: it receives realized future prices and loads over its planning horizon.

Each home's five-hour backup need $\hat N_{i,t}$ is the 90th percentile of positive net-load energy, divided by $\eta_d$, at nearby local-time slots in the 28 days preceding the episode. This is a usable-energy requirement; the optimization reserve target adds the physical minimum $e_i^{\min}$ before comparing it with battery state. Historical windows are included only if their complete energy requirement is observed before episode start. These reserve statistics remain fixed during the episode. Tier reserves use the same history and select each home's longest historical tier from 2, 4, 6, 8, 12, or 24 hours that fits 80\% of available capacity. If none fits, they fall back to the two-hour target clipped at 80\%, without claiming that tier is feasible. Full target-year profiles are not used to estimate controller reserves.

\paragraph{Outage-risk forecast.}
The \emph{hazard model} predicts whether a county's outage fraction will reach 1\% within the next $k$ hours, for $k\in\{1,3,6,12,24\}$. It is a county event forecast, not a calibrated probability that a particular home will fail. We fit one histogram gradient-boosted classifier per horizon on 2014--2019 data and calibrate its probabilities with isotonic regression on 2020. Both stages precede the 2021 extreme-event test.

Features comprise weather from completed hours, lagged outage fractions, issued alerts, and calendar variables. No realized future weather is used as a forecast proxy. Hourly model probabilities are carried forward to 15-minute dispatch intervals. For the optimization surrogate, nonnegative increments of the monotone cumulative horizon probabilities are spread across each horizon segment. These increments approximate the timing of county risk; they should not be interpreted as independently calibrated household outage-onset probabilities.

\section{Reserve policies and learned dispatch}
\label{sec:controllers}

All operating policies share the battery model, tariff, replay draws, and causal information described above. A reserve target defines how much energy is retained while grid-connected. Fixed reserves and alert rules are useful reference policies because their behavior is easy to inspect. Weather-conditioned charging also has precedents at a home or microgrid~\citep{schmitz2020proactive,gutierrezrojas2021weather}; related storage work learns terminal state-of-charge targets for arbitrage~\citep{tapia2026reachability}. Our empirical question is how these choices behave together in a residential fleet replay with outages and market shift.

\paragraph{Optimization family.}
The model-based controllers solve an independent finite-horizon dynamic program for each home over 32 energy levels and a 24-hour planning horizon. They replan hourly and execute four 15-minute actions. They include fixed 10\%, 20\%, 30\%, and 50\% targets; a 20\% floor with a soft preference for 90\% charge; an alert rule that targets full charge; and a historical quantile tier reserve inspired by \citet{anunrojwong2026pooling}. This implementation does not reproduce that paper's pooling optimization.

The \emph{risk-priced} controller retains a common 10\% operating minimum and adds
\begin{equation}
 \lambda_r\eta_d\sum_h \hat p_h\left(\hat N_{i,t+h}+e_i^{\min}-e_{i,t+h+1}\right)^+
 \label{eq:surrogate}
\end{equation}
to its cost, where $h=0,\ldots,H-1$, $H\le96$, and $\hat p_h$ is the county-risk timing proxy. The shortfall uses energy after the planned action in each interval. The \emph{hazard-threshold reserve} also retains the 10\% operating minimum. It raises the target to the estimated five-hour need plus the physical minimum when cumulative forecast risk reaches $\varepsilon\in\{0.02,0.1\}$, capped at 90\% capacity. Although motivated by chance constraints, this threshold rule does not establish an outage-service probability guarantee. An oracle-onset variant replaces forecast timing with realized future county threshold crossings. The separate market-lookahead reference uses perfect prices and loads, a 5\% physical-minimum target, and no future outage knowledge. Its discretized, rolling-horizon solution is a reference for market opportunity, not a proved global upper bound.

\paragraph{Learned policy.}
PPO outputs an aggregate power fraction and a reserve multiplier. The multiplier scales $\hat N_{i,t}$, clipped between 10\% and 90\% of available capacity. A priority allocator dispatches homes according to their energy above reserve, with individual inverter and energy limits~\citep{elsaadany2026realizability}. When a target rises above current energy, feasible mandatory charging takes precedence over the aggregate proposal. The layer enforces physical constraints and records recovery shortfall separately from violations of an attainable target. Such a layer does not establish outage reliability, which depends on the selected reserve and subsequent demand~\citep{stanojev2023safe,chen2021prof,ceusters2023safe,markgraf2025projection}.

The actor and critic each use two 256-unit tanh layers. The 39-dimensional observation includes local-time features, four lagged real-time prices, twelve publication-aware price forecasts, fleet SOC summaries, available power, historical need, alerts, current outage share, completed-hour weather, and five hazard forecasts. Gaussian actions are clipped to their allowed ranges. The policy uses summary statistics and a common reserve multiplier; it contains no set encoder or quadratic-program layer.

\section{Evaluation protocol}
\label{sec:protocol}

PPO training draws two-week episodes from Houston in 2019--2023. Fleet size varies from 50 to 200, hardware presets are randomized, state of health lies in $[0.85,1]$, and per-interval communication dropout ranges from zero to 10\%. Half of proposed episodes are sampled near recorded county events. Episodes overlapping named event holdouts are rejected. Training uses approximately three million transitions per run, 16 environments, rollout length 256, six optimization epochs, minibatches of 1,024, discount 0.997, GAE parameter 0.95, clipping 0.2, and initial learning rate $3\times10^{-4}$. Promise-cost runs use entropy coefficient 0.005. Appendix~\ref{app:repro} specifies the complete seed and penalty grid.

Validation uses Houston 2024, and chronological tests use Houston 2025 and available 2026 prices. Additional evaluations cover Dallas--Fort Worth, San Antonio, fleet size, hardware mix, and dropout. PPO's geographic transfer reuses a county-risk model already fitted on the twelve historical counties, including the six evaluation counties. Hardware and tariffs remain fixed across years, making historical operation a counterfactual replay. The 2026 bundle ends on August 15. Its target-year outage field is zero because the source outage record used here ends in 2025; it supports market-transfer comparisons only, with no 2026 reliability score. Changes across ERCOT's Real-Time Co-optimization Plus Batteries (RTC+B) redesign are temporal associations, not an identified causal effect of that redesign~\citep{ercot_rtcb_2025}.

Regular evaluation windows partition each available year into nonoverlapping intervals of at most 14 days. A preceding-history prefix supports January evaluation when the previous year is available; 2019 starts after its first 28 days. Named event windows are reported separately and never added again to regular-year averages. All controllers use the same sampled 100 homes, hardware mix, initial 90\% charge, and outage random numbers. Energy resets at each window boundary, so annualized statistics summarize episodic replays rather than continuous-year battery operation; interruptions touching a boundary are censored by that window. Larger fleets resample from the same 300-profile metropolitan pool, so fleet-size transfer does not establish transfer to thousands of distinct measured households.

Reported margin is Equation~\eqref{eq:margin}, plus a terminal inventory adjustment of $0.03\eta_d(e_{i,T}-e_{i,0})$ dollars. The same adjustment appears in the learned terminal reward. It limits gains from consuming initial inventory without assigning a speculative spike price to remaining charge. Annualized margins scale observed per-home cash and inventory value by observed duration; they are replay rates, not realized company annual profits. Both cash-only and adjusted metrics are retained in the release.

We compute promise rates from total failures divided by total episodes, rather than averaging window percentages. Margin differences are paired by window. Bootstrap intervals resample paired regular windows; reliability intervals recompute the ratio from resampled episode and failure counts. These intervals describe variation across the recorded windows under the fixed replay draw. They do not quantify uncertainty in the outage simulator, future market prices, or the source reporting coverage. We also report unserved energy, low-tail daily cash margins, per-home outcomes, physical violations, and dispersion across training seeds.

\section{An alert charging override and its limits}
\label{sec:guardrail}

The guardrail raises a home's target to $R_i=0.9E_i$ while a qualifying alert is active in its county. It explicitly overrides an idle or discharging proposal when charging is needed. This distinction matters: merely prohibiting further discharge cannot refill an already depleted battery. The override uses the same inverter, capacity, and availability limits as other dispatch.

\begin{proposition}[Conditional backup sufficiency]
\label{prop:charging}
Suppose an alert remains active throughout a grid-connected lead period of $n$ dispatch intervals before an outage. Home $i$ is available during that period, starts with energy $e_{i,a}$, and has fixed capacity and efficiencies. If
\begin{equation}
 n\Delta\ge \Delta\left\lceil\frac{(R_i-e_{i,a})^+}{\Delta\eta_c\bar P_i^c}\right\rceil,
\end{equation}
the override leaves at least $R_i$ energy at outage onset, independently of the proposed trading actions. If positive critical net demand never exceeds $\bar P_i^d$ and its first-five-hour energy requirement is no greater than $\eta_d(R_i-e_i^{\min})$, the five-hour promise is fulfilled.
\end{proposition}

\begin{proof}
Below $R_i$, the override charges at the lesser of the charge limit and the power needed to reach $R_i$ in one step. Grid connection and availability make that power feasible. The stated number of steps therefore reaches $R_i$; subsequent connected dispatch cannot discharge below it while the alert continues. After islanding, the stated pointwise power and cumulative energy conditions suffice to serve every positive demand increment before restoration or the five-hour endpoint. PV surplus can only add available energy.
\end{proof}

From the physical minimum, the conservative lead requirement is two hours for the 25 kWh and 39.2 kWh presets and four hours for the 50 kWh preset at full state of health. Short warnings, missing alerts, unavailable charging, and insufficient usable capacity violate the proposition's premises. Observed alert coverage and synthetic stress tests quantify these limitations below.

We make no bound on total margin loss using alert duty cycle alone. An alert changes stored energy, subsequent observations, and later dispatch, so two policies can continue to differ after it ends. Guardrail costs are measured over complete evaluation episodes, including those later differences. Low measured cost in one year is not a worst-case consistency guarantee.

\section{Results}
\label{sec:results}

\subsection{Reserve economics and backup service}

\begin{figure}[t]
  \centering
  \includegraphics[width=\linewidth]{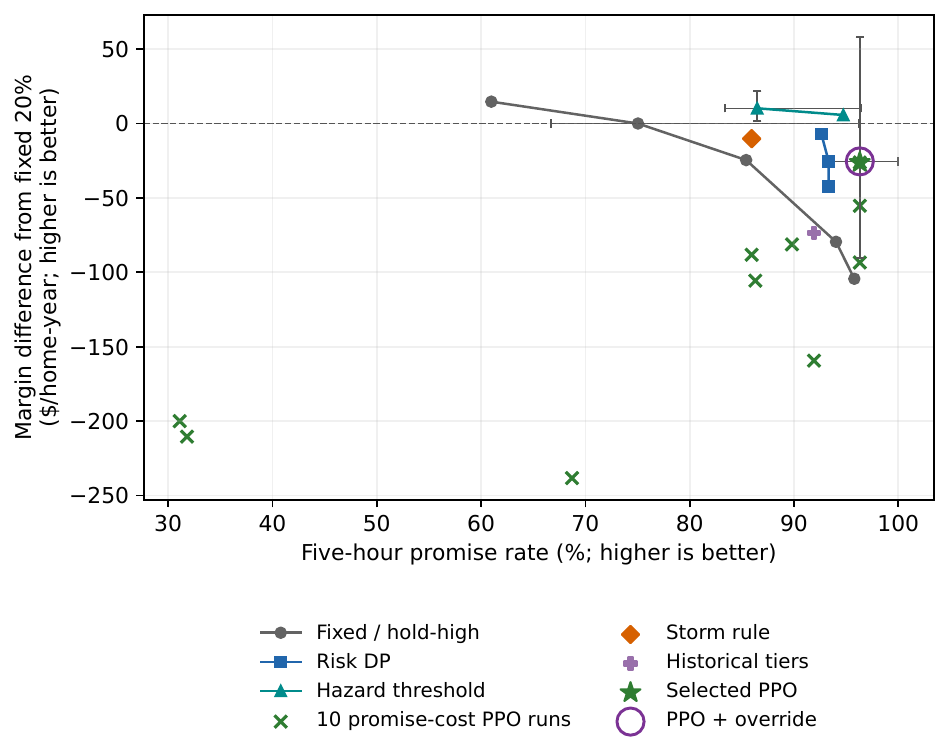}
  \caption{Houston 2024: economics versus simulated backup service. Higher values on either axis are preferable. Margin is paired against fixed 20\% and adjusted for terminal inventory. Crosses show every promise-cost PPO candidate; the selected candidate and its alert override are marked separately. Error bars for fixed 20\%, threshold 0.1, and guarded PPO show 95\% window-bootstrap intervals. Lines connect settings within a family and do not imply a fitted Pareto frontier.}
  \label{fig:frontier}
\end{figure}

Figure~\ref{fig:frontier} and Table~\ref{tab:frontier} report the Houston 2024 validation replay. A fixed 20\% reserve serves the first five hours of \FixedPromise\ of simulated episodes. The hazard-threshold rule with $\varepsilon=0.1$ yields \ThresholdPromise, with an annualized margin difference of \$\ThresholdDelta\ per home and paired 95\% interval $[\ThresholdLow,\ThresholdHigh]$. Its annualized unserved energy is \ThresholdUnserved\ kWh per home, compared with \FixedUnserved\ for fixed 20\%. Promise fulfillment and total unserved energy measure different parts of service: a controller can protect a short commitment while still exhausting its battery during a longer interruption.

The observed promise increase is \ThresholdPromiseDelta\ percentage points. Its paired window-bootstrap interval is $[\ThresholdPromiseDeltaLow,\ThresholdPromiseDeltaHigh]$ points and includes zero. Thus the simulated economic gain has a positive paired interval, while the service improvement is sensitive to which recorded storm windows appear in a resample. Neither interval establishes a future household reliability guarantee.

Tables~\ref{tab:tails}, \ref{tab:households}, and \ref{tab:seasonal} give cash tails, household outcomes, and seasonal window cohorts. The seasonal grouping retains each complete replay window and states how boundary-crossing windows are assigned; it does not change the full-year comparison or selection criterion.

\begin{table}[t]
  \centering\small
  \begin{tabular}{lrrr}
\toprule
Controller & $\Delta$ margin [95\% CI] & Promise [95\% CI] & Unserved \\
\midrule
Fixed 20\% & +0.0 [0.0, 0.0] & 75.0 [66.7, 96.3] & 204.0 \\
Fixed 50\% & -79.5 [-138.5, -34.1] & 94.0 [92.0, 100.0] & 189.1 \\
Hold near 90\% & -104.3 [-145.1, -72.6] & 95.8 [93.5, 100.0] & 177.8 \\
Storm rule & -10.2 [-17.6, -3.7] & 85.9 [83.3, 97.9] & 190.1 \\
Historical tiers & -73.6 [-126.4, -32.4] & 91.9 [90.3, 100.0] & 189.9 \\
Risk DP 5 & -25.4 [-37.7, -15.2] & 93.3 [87.0, 100.0] & 191.8 \\
Threshold 0.1 & +10.2 [1.7, 21.7] & 86.5 [83.4, 96.5] & 195.2 \\
Selected PPO & -26.1 [-91.2, 57.9] & 96.3 [93.5, 100.0] & 172.7 \\
PPO + override & -25.5 [-90.3, 58.1] & 96.3 [93.5, 100.0] & 172.7 \\
\bottomrule
\end{tabular}

  \caption{Houston 2024, 100 simulated homes. Margin difference is USD/home-year and unserved energy is kWh/home-year; promise is a percentage. Brackets are 95\% bootstrap intervals. Higher margin/promise and lower unserved energy are preferable. All regular windows, including the final partial window, receive duration-appropriate weight. Named event evaluations are excluded from these aggregates.}
  \label{tab:frontier}
\end{table}

The risk-priced dynamic program at $\lambda_r=5$ yields \RiskPromise\ promise and a \$\RiskDelta\ margin difference from fixed 20\%. This comparison exposes the cost of preserving energy under its surrogate risk charge. The released summary also includes all fixed floors, the high-SOC preference, historical tiers, both risk thresholds, and the oracle-onset timing reference. A difference between forecast and oracle-onset variants reflects this particular surrogate's response to changed inputs; it is not an identified fraction of the value of perfect information. The market-lookahead reference additionally changes price and load information, so it is reported separately.

\subsection{Learning, seed dispersion, and transfer}

The validation criterion selects \SelectedRun\ from ten promise-cost candidates. Its promise rate is \LearnedPromise\ and its margin difference is \$\LearnedDelta\ per home-year. Across those ten runs, promise spans \PromiseSeedLow--\PromiseSeedHigh\ and median margin difference is \$\PromiseSeedMedianDelta. Of these \PromiseCandidateCount\ runs, \PromiseBelowFixedCount\ have lower promise than fixed 20\%, and \PromiseNegativeMarginCount\ have lower inventory-adjusted margin. Table~\ref{tab:promise-seeds} reports every candidate, and Table~\ref{tab:other-seeds} reports all eleven full-cost and reserve-ablation runs. Mean SOC and throughput help distinguish retaining charge from actively timing market opportunities; a single selected point cannot describe training stability.

\begin{figure}[tbp]
  \centering
  \includegraphics[width=\linewidth]{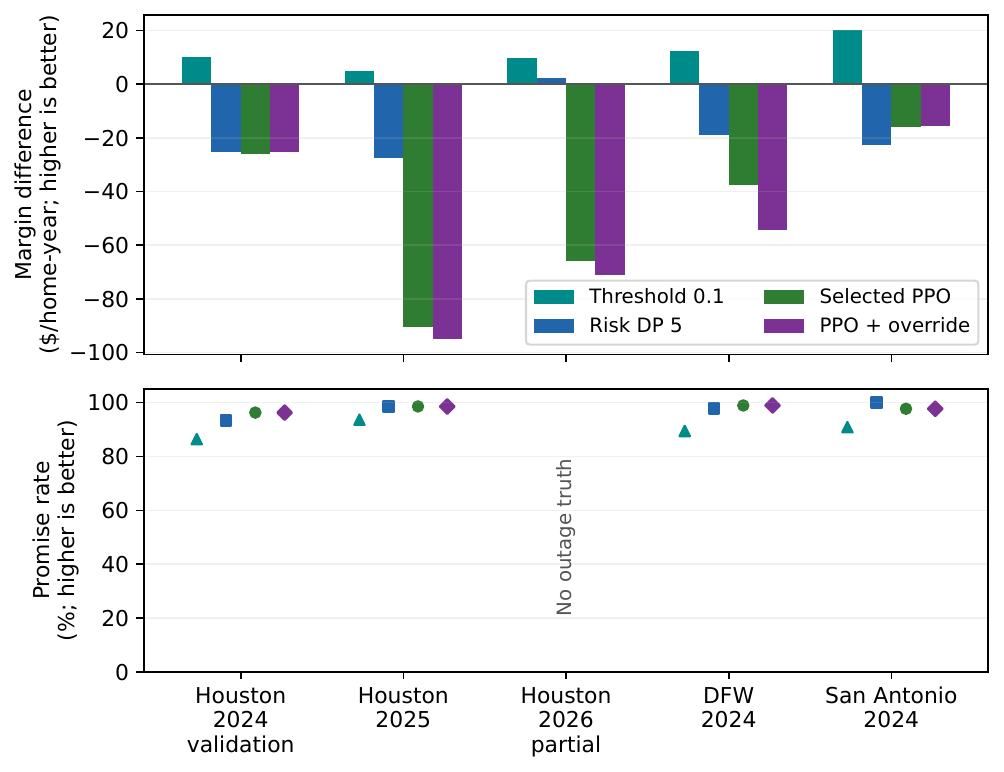}
  \caption{Transfer of the fixed validation choices. The upper panel shows annualized margin differences from the local fixed-20\% reference; positive is preferable. The lower panel shows episode-pooled five-hour promise; higher is preferable. The 2026 replay is partial-year and has no observed outage truth. Metro comparisons use the same tariff and hardware assumptions with different public building pools and county traces.}
  \label{fig:transfer}
\end{figure}

On Houston 2025, threshold 0.1 differs from fixed 20\% by \$\TestThresholdDelta\ per home-year, and the selected learned policy by \$\TestLearnedDelta; their promise rates are \TestThresholdPromise\ and \TestLearnedPromise. On available 2026 prices, the corresponding margin differences are \$\PartialThresholdDelta\ and \$\PartialLearnedDelta. These chronological tests assess reuse of the frozen controller. They do not retune a policy for the new year or infer the causal effect of a market redesign. Geographic outcomes appear in Figure~\ref{fig:transfer} and the complete released summary.

The validation improvement therefore does not uniformly transfer as a gain on both axes. In 2025, threshold 0.1 has higher modeled margin but lower promise than fixed 20\% (\TestFixedPromise). The market-only 2026 comparison cannot resolve that service tradeoff.

The threshold's 2024 metro outcomes also differ. In DFW it yields \DFWThresholdPromise\ promise versus \DFWFixedPromise\ for fixed 20\%, with a \$\DFWThresholdDelta\ margin difference per home-year. In San Antonio the corresponding values are \SanAntonioThresholdPromise\ versus \SanAntonioFixedPromise\ and \$\SanAntonioThresholdDelta. These comparisons pool \DFWEpisodes\ and \SanAntonioEpisodes\ simulated household episodes, respectively, with outcomes correlated through shared county traces.

Six pre-specified windows also test fleets of 100, 500, and 1,000 homes, a 70\% Core mix, and 20\% communication dropout (Table~\ref{tab:structure}). All five scenarios retain the same six windows and include fixed, storm, risk-DP, learned, and guarded learned controls. Those results describe sensitivity within the simulated pool. Resampling a larger fleet does not create new measured homes or establish asymptotic reliability. Table~\ref{tab:latency} reports complete CPU decision/simulation wall time under a stated backend; it is not a field communication-latency measurement.

\subsection{Storms, warning coverage, and the charging override}

Uri's scored intervals are held out of PPO training; it is an event holdout rather than a chronological backtest (Appendix~\ref{app:repro}). On its Houston window, selected PPO has promise \UriLearnedPromise\ and margin \$\UriLearnedMargin\ per home-week. Adding the alert override changes these values to \UriGuardedPromise\ and \$\UriGuardedMargin. Fixed 20\% and risk DP 5 yield \UriFixedPromise\ and \UriRiskPromise. Figure~\ref{fig:traces} shows energy and unserved-demand trajectories; Table~\ref{tab:events} also reports Elliott, derecho, Beryl, and a DFW Elliott replay. A zero-episode window has no reliability score. These lower-severity cases and the synthetic tests extend the audit beyond one extreme storm.

The warning system limits any interpretation of Proposition~\ref{prop:charging}. Across all six evaluated counties in 2019--2025, \MassEvents\ county episodes reach 5\% customers out. A qualifying alert is active at onset in \AlertCoverage, but continuous lead is at least four hours in only \AlertLeadCoverage\ (Table~\ref{tab:alerts}). Four hours is the conservative minimum-to-90\% charging time across all three hardware presets. Alert presence sometime during the preceding day is reported separately and is not substituted for an active, sustained warning. This audit uses county episodes merged across gaps of up to four hours, distinct from the one-hour household flicker merge in promise scoring.

\begin{figure}[tbp]
  \centering
  \includegraphics[width=\linewidth]{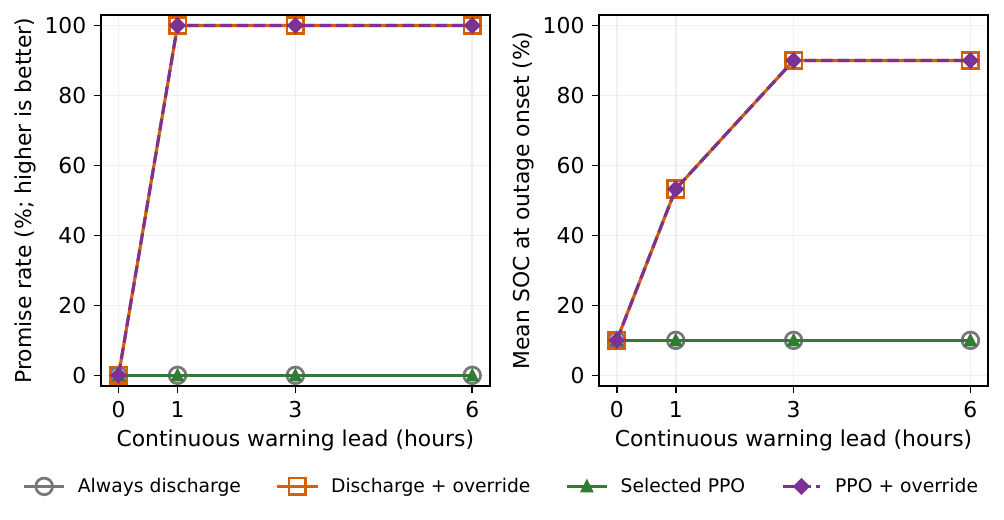}
  \caption{Controlled joint-event stress: \$5,000/MWh price, eight-hour outage, 10\% initial SOC, 25 kWh batteries, 2 kW demand, no PV/dropout, and full state of health. Warning lead varies while other inputs remain fixed. Promise rate is higher-is-better; onset SOC explains the available backup energy. The wider synthetic suite includes 86 indexed cases, including false/missed alerts, load uplift, reduced state of health, and dropout.}
  \label{fig:stress}
\end{figure}

The synthetic suite evaluates an always-discharge proposal and the selected PPO, each with and without the override. Every one of \SufficientTests\ guarded comparisons meeting the nominal sufficient-condition subset serves the five-hour promise. This verifies the implementation for those cases rather than extending the proposition to cases outside its premises. Short or missing warnings and unavailable charging require separate outcomes, which remain in the released results. False alerts can cost up to \$\FalseAlertWorstCost\ per home over a complete 48-hour synthetic episode. Their paired cash-margin differences after the warning ends reach \$\FalseAlertAfterDifference\ in magnitude. Consequently, charging cost cannot be assessed solely during alert hours.

\FloatBarrier
\section{Distribution and battery-life consequences}
\label{sec:system}

\paragraph{Static AC audit.}
Dispatch affects household net exchange even though market prices are exogenous. We replay three days on one SMART-DS Austin feeder~\citep{smartds}, randomly attach batteries at 30\% or 100\% of load locations, and compare with identical load/PV profiles and no battery dispatch. This is an illustrative placement study using Houston profiles on a synthetic Austin topology, not a calibrated network representation of those homes. Grid exchange is zero while a simulated home is islanded. The original feeder load models are retained, with reactive demand set to 0.33 times positive real demand.

Table~\ref{tab:ac} reports convergence, voltage ranges, and the fraction of intervals with voltage or line-rating violations. The aggregate solver-convergence rate is \ACConvergence. Under the stated reporting rule, the reference already exhibits undervoltage, overvoltage, and line overloads; 30\% penetration cannot be described as uniformly benign. We retain every phase-node voltage at buses reachable in the bus-level source graph, including near-zero values, rather than filtering by an acceptable-voltage band. Eighteen phase nodes at unreachable buses are reported separately. Bus-level reachability does not certify phase energization. Snapshot convergence alone is not proof of acceptable operation, protection coordination, or dynamic stability.

\paragraph{Aging-model sensitivity.}
Table~\ref{tab:aging} compares four windows and 20 homes per window. A throughput model predicts \FixedETFade\ annualized fade for fixed 20\% and \HoldETFade\ for the hold-high preference. Under the SOC/temperature-dependent BLAST model, repeating those windows predicts \FixedBLASTFade\ and \HoldBLASTFade, respectively. These comparisons expose different penalties for cycling and high SOC. They do not measure the lifetime of the specified commercial packs. The throughput coefficient uses a complete charge/discharge cycle as its normalization; the corrected release records the factor-of-two reporting correction to earlier outputs. The linear-calendar, throughput, and rainflow coefficients and the BLAST chemistry/repetition convention are specified in Appendix~\ref{app:aging}. Dispatch is held fixed across aging models and receives no aging charge in the main comparison.

\section{Discussion and limits}

This replay supports comparisons conditional on its input records and simulator. It does not estimate the reliability of an operating battery company. County outage fractions cannot identify which sampled buildings failed, missing county reports are treated as zero, and a repeated 2018 building profile does not represent storm-specific heating behavior. Synthetic load uplift tests one consequence of that mismatch without reconstructing historical Uri demand. The fixed modeled-customer denominator and evolving reporting coverage add further uncertainty. Warning mapping uses a single April 2026 zone/county crosswalk across historical years, so boundary changes can affect the coverage audit.

The replay's input lags define assumed availability, rather than verify every historical publication vintage. Weather comes from retrospective model/reanalysis products, and warning intervals use archived issue/expiry fields rather than every as-issued message. Actual reporting delays, revisions, and operational weather feeds remain untested.

The reliability budget is a useful statement of the desired service, but a shared fixed penalty and a common learned reserve multiplier do not enforce it for every household. The learned action parameterization, fleet summary observations, one PPO implementation, and finite training budget restrict the conclusions to this controller family. A failure here does not establish that all learned reserve policies must fail. Conversely, selecting one candidate from ten on a validation year does not remove selection uncertainty; all candidate outcomes remain visible.

Economic comparisons omit capital, hedging, subscriptions, ancillary-service revenues and performance obligations, and aging costs. Repeated two-week initializations can support different behavior from continuous operation, even with a terminal inventory adjustment. Bootstrap intervals reflect recorded-window variation under one replay seed, rather than uncertainty in the household outage model or future event frequencies. Prices after the RTC+B redesign provide a chronological stress test, not an estimate of the redesign's causal effect.

The alert proposition states conditions that can be checked in a specified model. Real warning lead and charging availability often fail those conditions. Neither favorable replay outcomes nor a conditional proposition validates unattended deployment. The feeder audit likewise shows why individual battery feasibility is insufficient to establish network acceptability: loads, placement, voltage behavior, thermal ratings, and protection need their own calibrated assessment. Continuous-year replay, household outage telemetry, operational weather forecasts, household-specific risk enforcement, and network-aware control would address distinct limitations left open here.

\section*{Data and code availability}

The public repository is \url{https://github.com/MaruthiV/homefleet}. Release v0.2.0 contains code and hash-identified data, checkpoint, and result archives. A packaged eight-home example runs without a download. The data card distinguishes observations from simulations and records source terms; the reproducibility guide maps commands to each result. The named preprint uses protocol \texttt{causal-v3}; its outputs should not be pooled with the original workshop submission. The repository records the substantive scientific corrections.

\section*{LLM usage disclosure}

Language models assisted with implementation, analysis scripts, source checking, and prose editing. The released code, assumptions, input manifests, numerical outputs, and manuscript-generation scripts make those contributions inspectable.

\FloatBarrier
\bibliographystyle{plainnat}
\setlength{\bibsep}{2pt}
\bibliography{refs}

@article{anunrojwong2026pooling,
  title={Residential Battery Pooling Under Backup Commitments},
  author={Anunrojwong, Jerry and Zhang, Baosen},
  journal={arXiv preprint arXiv:2605.17723},
  year={2026},
  url={https://arxiv.org/abs/2605.17723}
}

@article{humiston2026degradation,
  title={Evaluating Battery Degradation Models in Rolling-Horizon {BESS} Arbitrage Optimization},
  author={Humiston, Chase and Cetin, Mehmet and de Queiroz, Anderson Rodrigo},
  journal={Energies},
  volume={19},
  number={4},
  pages={1056},
  year={2026},
  doi={10.3390/en19041056}
}

@article{schmitz2020proactive,
  title={Proactive home energy storage management system to severe weather scenarios},
  author={Schmitz, William Ismael and Schmitz, Magdiel and Canha, Luciane Neves and Garcia, Vin{\'i}cius Jacques},
  journal={Applied Energy},
  volume={279},
  pages={115797},
  year={2020},
  doi={10.1016/j.apenergy.2020.115797}
}

@article{gutierrezrojas2021weather,
  title={Weather-Driven Predictive Control of a Battery Storage for Improved Microgrid Resilience},
  author={Guti{\'e}rrez-Rojas, Daniel and Mashlakov, Aleksei and Brester, Christina and Niska, Harri and Kolehmainen, Mikko and Narayanan, Arun and Honkapuro, Samuli and Nardelli, Pedro H. J.},
  journal={IEEE Access},
  volume={9},
  pages={163108--163121},
  year={2021},
  doi={10.1109/ACCESS.2021.3133490}
}

@article{gorman2025backup,
  title={Backup power or bill savings? {How} electricity tariffs impact residential solar-plus-storage usage in the {United States}},
  author={Gorman, Will and Barbose, Galen and Baik, Sunhee and Miller, Cesca and Carvallo, Juan Pablo},
  journal={Utilities Policy},
  volume={94},
  pages={102035},
  year={2025},
  doi={10.1016/j.jup.2025.102035}
}

@article{tapia2026reachability,
  title={Learning Reachability of Energy Storage Arbitrage},
  author={Tapia, Tom{\'a}s and Castellano, Agustin and Mallada, Enrique and Dvorkin, Yury},
  journal={arXiv preprint arXiv:2512.06600},
  year={2026},
  url={https://arxiv.org/abs/2512.06600}
}

@inproceedings{stanojev2023safe,
  title={Safe Reinforcement Learning for Strategic Bidding of Virtual Power Plants in Day-Ahead Markets},
  author={Stanojev, Ognjen and Mitridati, Lesia and de Nardis di Prata, Riccardo and Hug, Gabriela},
  booktitle={IEEE SmartGridComm},
  year={2023},
  url={https://arxiv.org/abs/2307.05812}
}

@inproceedings{chen2021prof,
  title={Enforcing Policy Feasibility Constraints through Differentiable Projection for Energy Optimization},
  author={Chen, Bingqing and Donti, Priya L. and Baker, Kyri and Kolter, J. Zico and Berg{\'e}s, Mario},
  booktitle={ACM e-Energy},
  year={2021},
  url={https://arxiv.org/abs/2105.08881}
}

@article{ceusters2023safe,
  title={An adaptive safety layer with hard constraints for safe reinforcement learning in multi-energy management systems},
  author={Ceusters, Glenn and Putratama, Muhammad Andy and Franke, R{\"u}diger and Now{\'e}, Ann and Messagie, Maarten},
  journal={Sustainable Energy, Grids and Networks},
  year={2023},
  url={https://arxiv.org/abs/2304.08897}
}

@article{markgraf2025projection,
  title={Safe Reinforcement Learning using Action Projection: Safeguard the Policy or the Environment?},
  author={Markgraf, Hannah and Sawant, Shambhuraj and Krasowski, Hanna and Sch{\"a}fer, Lukas and Gros, S{\'e}bastien and Althoff, Matthias},
  journal={arXiv preprint arXiv:2509.12833},
  year={2025},
  url={https://arxiv.org/abs/2509.12833}
}

@article{elsaadany2026realizability,
  title={Making Aggregations Reliable: Realizability Guarantees for Battery Fleets with Heterogeneous Power and Energy Limits},
  author={Elsaadany, Mazen and Almassalkhi, Mads R.},
  journal={arXiv preprint arXiv:2606.01562},
  year={2026},
  url={https://arxiv.org/abs/2606.01562}
}

@inproceedings{karimimadahi2024control,
  title={Control Policy Correction Framework for Reinforcement Learning-based Energy Arbitrage Strategies},
  author={Karimi Madahi, Seyed Soroush and Gokhale, Gargya and Verwee, Marie-Sophie and Claessens, Bert and Develder, Chris},
  booktitle={ACM e-Energy},
  year={2024},
  url={https://arxiv.org/abs/2404.18821}
}

@article{brelsford2024eaglei,
  title={A dataset of recorded electricity outages by {United States} county 2014--2022},
  author={Brelsford, Christa and others},
  journal={Scientific Data},
  volume={11},
  pages={271},
  year={2024},
  doi={10.1038/s41597-024-03095-5}
}

@misc{eaglei_v4,
  title={The {Environment} for {Analysis} of {Geo-Located} {Energy} {Information}'s {Recorded} {Electricity} {Outages} 2014--2025 (v4)},
  author={Brelsford, Christa and others},
  year={2026},
  howpublished={figshare},
  doi={10.6084/m9.figshare.24237376.v4},
  url={https://doi.org/10.6084/m9.figshare.24237376.v4}
}

@inproceedings{yeh2023sustaingym,
  title={SustainGym: Reinforcement Learning Environments for Sustainable Energy Systems},
  author={Yeh, Christopher and Li, Victor and Datta, Rajeev and Arroyo, Julio and Christianson, Nicolas and Zhang, Chi and Chen, Yize and Hosseini, Mohammad Mehdi and Golmohammadi, Azarang and Shi, Yuanyuan and Yue, Yisong and Wierman, Adam},
  booktitle={NeurIPS Datasets and Benchmarks},
  year={2023},
  url={https://papers.neurips.cc/paper_files/paper/2023/file/ba74855789913e5ed36f87288af79e5b-Paper-Datasets_and_Benchmarks.pdf}
}

@article{schulman2017ppo,
  title={Proximal Policy Optimization Algorithms},
  author={Schulman, John and Wolski, Filip and Dhariwal, Prafulla and Radford, Alec and Klimov, Oleg},
  journal={arXiv preprint arXiv:1707.06347},
  year={2017},
  url={https://arxiv.org/abs/1707.06347}
}

@inproceedings{lin2025accuracy,
  title={Maximizing the Value of Predictions in Control: Accuracy Is Not Enough},
  author={Lin, Yiheng and Yeh, Christopher and Chen, Zaiwei and Wierman, Adam},
  booktitle={NeurIPS},
  year={2025},
  url={https://arxiv.org/abs/2506.04497}
}

@article{xu2018rainflow,
  title={Factoring the Cycle Aging Cost of Batteries Participating in Electricity Markets},
  author={Xu, Bolun and Zhao, Jinye and Zheng, Tongxin and Litvinov, Eugene and Kirschen, Daniel S.},
  journal={IEEE Transactions on Power Systems},
  volume={33},
  number={2},
  pages={2248--2259},
  year={2018},
  doi={10.1109/TPWRS.2017.2733339}
}

@misc{blastlite,
  title={{BLAST-Lite}: Battery Lifetime Analysis and Simulation Toolsuite},
  author={Gasper, Paul and Prakash, Nina and Smith, Kandler},
  howpublished={\url{https://github.com/NatLabRockies/BLAST-Lite}},
  note={Version 1.1.1; accessed November 2026},
  year={2026}
}

@misc{resstock2024,
  title={{ResStock} End-Use Load Profiles for the {U.S.} Building Stock, 2024 {AMY2018} Release 2},
  author={{NREL}},
  howpublished={Open Energy Data Initiative},
  year={2024},
  url={https://data.openei.org/submissions/4520}
}

@misc{teslastormwatch,
  title={Powerwall Storm Watch},
  author={{Tesla, Inc.}},
  howpublished={\url{https://www.tesla.com/support/energy/powerwall/mobile-app/storm-watch}},
  note={Accessed November 2026},
  year={2026}
}

@misc{basecharge2026,
  title={Understanding how {Base} charges and discharges its batteries},
  author={{Base Power Company}},
  howpublished={\url{https://www.basepowercompany.com/blog/how-base-charges-and-discharges-its-batteries}},
  note={Accessed November 2026},
  year={2026}
}

@techreport{ice2025,
  title={{ICE} Calculator 2.0: Final Report for Phase 1 of the National Initiative to Update the Interruption Cost Estimate ({ICE}) Calculator},
  author={Larsen, Peter and others},
  institution={Lawrence Berkeley National Laboratory},
  year={2025},
  doi={10.2172/2569414}
}

@article{brown2024reliability,
  title={The value of electricity reliability: Evidence from battery adoption},
  author={Brown, David P. and Muehlenbachs, Lucija},
  journal={Journal of Public Economics},
  volume={239},
  pages={105216},
  year={2024},
  doi={10.1016/j.jpubeco.2024.105216}
}

@misc{ercot_rtcb_2025,
  title={{ERCOT} Goes Live with Real-Time Co-optimization Plus Batteries ({RTC+B})},
  author={{ERCOT}},
  year={2025},
  howpublished={\url{https://www.ercot.com/news/release/12052025-ercot-goes-live}}
}

@misc{smartds,
  title={{SMART-DS}: Synthetic Models for Advanced, Realistic Testing: Distribution Systems and Scenarios},
  author={{NREL}},
  year={2020},
  howpublished={Open Energy Data Initiative},
  url={https://data.openei.org/submissions/2981}
}

@misc{openmeteo_archive,
  title={Historical Weather API},
  author={{Open-Meteo}},
  year={2026},
  howpublished={\url{https://open-meteo.com/en/docs/historical-weather-api}},
  note={Accessed November 11, 2026}
}

@misc{iem_vtec,
  title={{NWS} Valid Time Extent Code ({VTEC}) Archives},
  author={{Iowa Environmental Mesonet}},
  year={2026},
  howpublished={\url{https://mesonet.agron.iastate.edu/info/datasets/vtec.html}},
  note={Accessed November 11, 2026}
}
\clearpage
\appendix
\section{Notation and implementation details}
\label{app:notation}

\begin{table}[htbp]
\centering\small
\begin{tabular}{ll}
\toprule
Symbol & Meaning and units \\
\midrule
$i,c,z,t$ & Home, county, load zone, dispatch interval \\
$\Delta$ & Dispatch duration, 0.25 h \\
$L_{i,t},V_{i,t}$ & Household demand and PV output, kW \\
$\bar E_i,E_i$ & Rated and SOH-adjusted available capacity, kWh \\
$e_{i,t},e_i^{\min}$ & Stored energy and physical minimum, kWh \\
$R_{i,t}$ & Connected operating reserve target, kWh \\
$p^c_{i,t},p^d_{i,t}$ & Nonnegative charging/discharging power, kW \\
$m_{i,t},x_{i,t}$ & Connected import/export power, kW \\
$\lambda_{z,t}$ & Real-time settlement price, \$/kWh in accounting \\
$U_{i,t},B_i$ & Interval unserved energy and annual reliability budget, kWh \\
$\lambda_r$ & Shared reliability penalty, \$/kWh of delivered energy \\
$\hat N_{i,t}$ & Historical five-hour usable-energy need, kWh stored \\
$\hat p_h$ & Forecast county-risk timing proxy at planning step $h$ \\
$f_{c,t}$ & Recorded county customers-out fraction \\
$u_p,u_r$ & Learned power fraction and reserve multiplier \\
$\varepsilon$ & Cumulative-risk threshold in the planning horizon \\
\bottomrule
\end{tabular}
\caption{Notation used in the model and results. Stored-energy requirements account for discharge efficiency; delivered unserved energy is scored in kWh at the household.}
\end{table}

The three hardware presets are $(\bar E,\bar P^c,\bar P^d)=(25,11.4,11.4)$, $(50,11.4,11.4)$, and $(39.2,20,20)$ in kWh/kW/kW. The reference evaluation mix is 60\%, 20\%, and 20\%, respectively. Both efficiencies are 0.9486833. The physical minimum is 5\% of rated energy; operating percentage targets use available energy. State of health is fixed at one in the principal evaluation, and randomized during training. Homes are sampled with replacement from each metropolitan pool.

For the five-hour need, each historical sample sums positive demand net of PV over 20 intervals, divides by discharge efficiency, and is grouped by local-time slot. The 90th percentile pools the target slot and its two neighboring quarter-hour slots on either side. Only samples whose entire forward window ends before episode start are eligible. Tier requirements repeat the calculation at their respective durations. PV surplus does not reduce a later positive-demand requirement in this conservative statistic; it can charge the battery in actual replay.

The learned reserve mapping is
\begin{equation}
R_{i,t}=\operatorname{clip}\left(u_r\hat N_{i,t},\,0.1E_i,\,0.9E_i\right),\qquad 0\le u_r\le1.
\end{equation}
This is an action parameterization, not an exact translation of the optimization shortfall target $\hat N+e^{\min}$. In particular, outside the alert override its multiplier cannot request arbitrarily more than the historical usable-energy statistic. That restriction is part of the evaluated learned controller, and limits claims about other learned reserve designs. The alert override takes the maximum of this target and $0.9E_i$ for affected homes.

The power proposal is $u_p$ times the currently realizable aggregate charge or discharge envelope, according to its sign. The priority allocator respects each home's feasible envelope. If enforcing a raised target requires additional charging, actual aggregate power can differ from that proposal. The absolute gap receives a reward penalty of 0.01 dollars per kW before division by fleet size. This engineering penalty is separate from the unserved-energy multiplier. Communication dropout is sampled independently per interval/home and disables remote trading actions; autonomous local islanded service remains available.

\paragraph{Optimization reference.}
Each home uses 32 equally spaced energy states between its physical minimum and available capacity. The horizon is at most 96 quarter-hours and shrinks at episode end. The program replans every four intervals, using current measured energy snapped to the nearest grid state for planning. Actual dispatch subsequently passes through continuous physical clipping and mandatory target recovery. Its terminal value is energy times discharge efficiency times the mean forecast wholesale price over that planning horizon; this is distinct from the fixed terminal inventory value used to compare realized episode margins. The hold-near-90\% reference adds a 0.02 \$/kWh-hour penalty on its energy gap below 90\%, above a 20\% floor. It is a soft high-SOC preference rather than a hard 90\% minimum.

An operating-floor shortfall receives a large planning penalty, while the execution layer enforces the target to the extent physically reachable within the interval. Risk DP adds Equation~\eqref{eq:surrogate}; the threshold rule raises each future planning step's floor when cumulative timing-proxy mass through that step reaches $\varepsilon$. Neither anticipates future individual islanding. The oracle-onset variant substitutes realized county 1\% threshold onsets, rather than realized home failures. Oracle onset and market lookahead are two separate information references.

\section{Complete training and reproduction protocol}
\label{app:repro}

The frozen main grid contains 21 runs. Promise-window training uses $\lambda_r\in\{5,20\}$ and seeds 1--5. Full-outage-cost learned-reserve training uses $\lambda_r=1$ at seeds 0--1, $\lambda_r=5$ at seeds 0--2, and $\lambda_r=20$ at seeds 0--1. Fixed-reserve and physical-minimum-only learned trading ablations use $\lambda_r=5$ at seeds 0--1. All retain physical clipping and the same policy architecture. Thus the physical-minimum-only ablation is not an unconstrained battery simulator.

Each run uses 2,998,272 transitions, or 732 complete rollouts of 16 environments times 256 intervals. There are six PPO epochs and four minibatches per rollout, Adam epsilon $10^{-5}$, gradient-norm clipping 0.5, value-loss coefficient 0.5, and a linearly decaying learning rate from $3\times10^{-4}$. Discount, GAE, and policy-clipping parameters are 0.997, 0.95, and 0.2. Entropy coefficient is 0.005 for promise-cost runs and zero for full-cost runs. Observations and discounted returns have online running mean/variance normalization. Minibatch advantages are standardized. Gaussian samples are clipped into the environment's power/reserve action ranges; the log-probability objective uses the sampled Gaussian action. This clipped-action convention is a limitation of this implementation, not a newly proposed PPO estimator.

The reference run uses one ARM64 CPU workstation with 11 logical processors and 18 GiB memory, up to six concurrent training processes, and one PyTorch thread per run. No GPU is used. Per-run configuration and progress logs retain the exact library versions, transition counts, and training times. File locks serialize any overlapping requests for the same run; parallel scheduling changes wall time, not the per-run budget or seed.

Training samples Houston 2019--2023, fleet size uniformly from 50--200, hardware presets uniformly across the three types, per-home SOH uniformly from 0.85--1, and per-episode dropout probability uniformly from 0--0.1. Half of proposals start near county events above 1\%, with a random offset of up to three days. Every proposed interval overlapping a named training-year event holdout is rejected. A failed search raises an error rather than silently accepting a holdout overlap. Historical feature buffers after an event can contain its past observations; no additional 28-day feature embargo is imposed.

Storm tests dated before 2024 are event holdouts within the training-year range, not chronological deployment backtests: the controller also trains on other, later dates in 2019--2023. Houston 2025 and 2026 are the chronological tests that postdate the entire PPO training period. ``Causal'' in the data protocol refers to the decision-time information set, including the pre-Uri hazard fit/calibration, rather than to a claim that the PPO controller was trained before every named storm.

Among the ten promise-cost candidates, selection maximizes Houston 2024 annualized adjusted margin minus 20 dollars per kWh of annualized first-five-hour unserved energy. Ties use controller name. The frozen selection is \SelectedRun. Neither guarded outcomes nor test-year/event performance enter this rule. Both guarded and unguarded versions of every checkpoint are evaluated on each main dataset; the selected comparison does not hide the remaining runs.

The main evaluation has five regular-year/metro datasets: Houston 2024, 2025, and partial 2026; DFW 2024; and San Antonio 2024. Six additional datasets are event-only: Houston 2021, 2022, and 2023; DFW 2021 and 2022; and San Antonio 2021. The three 2024 regular datasets also include named events in their output files, but those rows are never used in annualized aggregates. A machine-readable manifest lists every expected window and controller across 473 result files, and verifies exact coverage.

Replay uses seed zero for fleet selection, hardware assignment, and the separately initialized outage/availability draws. The homes and outage realizations are common across controllers within a dataset/scenario. Regular windows start at the first target-year timestamp after the preceding-history prefix and partition the available period without overlap. The final partial interval is retained. All episodes start with 90\% available charge. Annualization divides complete per-home totals by observed hours and multiplies by 8,760, including for the leap year and partial 2026; these are standardized rates.

The paper's aggregated cash-tail statistics use complete 24-hour blocks anchored at each window start; partial blocks are excluded. Per-window diagnostic columns can include a final short block and are not averaged to obtain these aggregates. CVaR$_5$ averages the lowest 5\% of those daily cash margins under the released quantile convention. The worst seven-block sum can cross a reset boundary and is not a continuous-operation worst-week estimate. Per-home promise quantiles include only homes with a simulated episode, while unserved-energy quantiles include every sampled home. Bootstrap intervals use 2,000 paired regular-window resamples with seed zero, pool episode/failure counts in each resample, and omit reliability resamples with no episodes.

Code and processed artifacts carry protocol and source hashes. All reported replays use the final \texttt{reserve-minimum-v3} controller revision and the corrected full-cycle throughput-aging normalization. Preliminary outputs are archived separately. Corrections to the training information, reward, and physical execution required fresh PPO training; the later change to the independent optimization controller does not alter the learned training dynamics.

\begin{table}[htbp]
\centering\small
\begin{tabular}{lrrrr}
\toprule
Run & $\Delta$ margin & Promise (\%) & Mean SOC (\%) & Throughput \\
\midrule
pc\_lam20\_s1 & -93.4 & 96.3 & 98.7 & 1,106.4 \\
pc\_lam20\_s2 & -105.6 & 86.3 & 96.2 & 2,090.4 \\
pc\_lam20\_s3 & -81.3 & 89.8 & 89.7 & 2,797.2 \\
pc\_lam20\_s4 & -159.3 & 91.9 & 56.5 & 12,048.3 \\
pc\_lam20\_s5 & -200.0 & 31.1 & 10.9 & 1,084.3 \\
pc\_lam5\_s1 & -55.3 & 96.3 & 97.4 & 2,323.0 \\
pc\_lam5\_s2 & -88.2 & 85.9 & 93.7 & 4,197.8 \\
pc\_lam5\_s3 & -238.3 & 68.7 & 45.0 & 14,808.9 \\
pc\_lam5\_s4* & -26.1 & 96.3 & 93.6 & 3,566.0 \\
pc\_lam5\_s5 & -210.4 & 31.8 & 20.2 & 1,439.8 \\
\bottomrule
\end{tabular}

\caption{All ten promise-cost candidates on Houston 2024. Asterisk denotes the fixed validation selection. Margin difference is \$/home-year against fixed 20\%; throughput is charge plus discharge kWh/home-year. Promise and mean SOC are percentages. Names encode penalty and seed.}
\label{tab:promise-seeds}
\end{table}

\begin{table}[htbp]
\centering\small
\begin{tabular}{lrrrr}
\toprule
Run & $\Delta$ margin & Promise (\%) & Mean SOC (\%) & Throughput \\
\midrule
fixed\_lam5\_s0 & -100.3 & 94.7 & 65.4 & 2,502.2 \\
fixed\_lam5\_s1 & -86.3 & 96.3 & 97.0 & 754.6 \\
learned\_lam1\_s0 & -46.9 & 95.6 & 88.9 & 1,207.5 \\
learned\_lam1\_s1 & -89.9 & 90.2 & 92.6 & 4,377.0 \\
learned\_lam20\_s0 & -2.2 & 80.3 & 95.1 & 3,225.7 \\
learned\_lam20\_s1 & -83.8 & 96.3 & 98.4 & 1,298.9 \\
learned\_lam5\_s0 & -137.0 & 90.2 & 38.5 & 2,132.7 \\
learned\_lam5\_s1 & -72.8 & 96.3 & 97.8 & 2,089.0 \\
learned\_lam5\_s2 & -106.4 & 86.3 & 96.5 & 2,767.6 \\
none\_lam5\_s0 & -148.5 & 64.7 & 12.7 & 2,121.8 \\
none\_lam5\_s1 & -232.3 & 39.7 & 13.7 & 1,979.6 \\
\bottomrule
\end{tabular}

\caption{All eleven full-outage-cost and reserve-ablation runs, with the same units as Table~\ref{tab:promise-seeds}. ``None'' removes the operating reserve but retains physical battery limits. These runs do not participate in selecting the headline promise-cost candidate.}
\label{tab:other-seeds}
\end{table}
\FloatBarrier

\clearpage
\section{Additional economic and household outcomes}
\label{app:tails}
\begingroup
\setlength{\intextsep}{8pt}

\begin{table}[!htbp]
\centering\small
\begin{tabular}{lrrrr}
\toprule
Controller & Cash/yr & Adjusted/yr & Daily CVaR$_5$ & Worst 7 blocks \\
\midrule
Fixed 20\% & 493.0 & 475.3 & -0.47 & 1.30 \\
Fixed 50\% & 405.9 & 395.8 & -0.48 & 1.19 \\
Hold near 90\% & 373.9 & 371.0 & -0.79 & 0.56 \\
Storm rule & 482.0 & 465.2 & -0.91 & 0.83 \\
Historical tiers & 414.5 & 401.7 & -0.48 & 1.22 \\
Risk DP 5 & 467.7 & 449.9 & -0.55 & 1.11 \\
Threshold 0.1 & 505.2 & 485.6 & -0.61 & 1.32 \\
Selected PPO & 448.6 & 449.2 & -1.39 & -0.96 \\
PPO + override & 449.3 & 449.9 & -1.35 & -0.64 \\
\bottomrule
\end{tabular}

\caption{Houston 2024 gross outcomes, in dollars per home. Cash and adjusted margins are annualized; daily CVaR and worst seven-block values use complete 24-hour cash blocks (Appendix~\ref{app:repro}). Higher values are preferable. Aging charges are excluded.}
\label{tab:tails}
\end{table}

\begin{table}[!htbp]
\centering\small
\begin{tabular}{lrrr}
\toprule
Controller & Homes with outages & Promise $p_5$\% & Unserved $p_{95}$/yr \\
\midrule
Fixed 20\% & 100 & 33.3 & 577.4 \\
Fixed 50\% & 100 & 66.2 & 547.2 \\
Hold near 90\% & 100 & 66.7 & 526.6 \\
Storm rule & 100 & 42.7 & 547.7 \\
Historical tiers & 100 & 49.6 & 548.4 \\
Risk DP 5 & 100 & 59.9 & 559.2 \\
Threshold 0.1 & 100 & 50.0 & 566.3 \\
Selected PPO & 100 & 74.8 & 518.0 \\
PPO + override & 100 & 74.8 & 518.0 \\
\bottomrule
\end{tabular}

\caption{Houston 2024 household tails. Promise $p_5$ is the fifth percentile of household episode-success rates among homes with outages (higher is better). Unserved $p_{95}$ is the 95th percentile of annualized household unserved energy, kWh/home-year (lower is better). These empirical quantiles do not certify household risk budgets.}
\label{tab:households}
\end{table}

\begin{table}[!htbp]
\centering\small
\begin{tabular}{llrrrr}
\toprule
Season & Dispatch & Hours & $\Delta$ margin/wk & Promise\% & Episodes \\
\midrule
DJF & Fixed 20\% & 2,400 & +0.0 & 88.9 & 18 \\
DJF & Threshold 0.1 & 2,400 & +0.0 & 88.9 & 18 \\
DJF & Risk DP 5 & 2,400 & -0.3 & 100.0 & 18 \\
DJF & Selected PPO & 2,400 & -0.7 & 100.0 & 18 \\
DJF & PPO + override & 2,400 & -0.6 & 100.0 & 18 \\
MAM & Fixed 20\% & 2,016 & +0.0 & 83.1 & 219 \\
MAM & Threshold 0.1 & 2,016 & +0.4 & 88.1 & 219 \\
MAM & Risk DP 5 & 2,016 & -0.3 & 87.2 & 219 \\
MAM & Selected PPO & 2,016 & -2.0 & 93.2 & 219 \\
MAM & PPO + override & 2,016 & -2.0 & 93.2 & 219 \\
JJA & Fixed 20\% & 2,352 & +0.0 & 67.4 & 316 \\
JJA & Threshold 0.1 & 2,352 & +0.1 & 84.5 & 316 \\
JJA & Risk DP 5 & 2,352 & -0.9 & 96.8 & 316 \\
JJA & Selected PPO & 2,352 & -0.4 & 98.1 & 316 \\
JJA & PPO + override & 2,352 & -0.4 & 98.1 & 316 \\
SON & Fixed 20\% & 2,016 & +0.0 & 100.0 & 16 \\
SON & Threshold 0.1 & 2,016 & +0.3 & 100.0 & 16 \\
SON & Risk DP 5 & 2,016 & -0.5 & 100.0 & 16 \\
SON & Selected PPO & 2,016 & +1.1 & 100.0 & 16 \\
SON & PPO + override & 2,016 & +1.1 & 100.0 & 16 \\
\bottomrule
\end{tabular}

\caption{Houston 2024 seasonal window cohorts, assigned by first local timestamp: DJF (Dec--Feb), MAM (Mar--May), JJA (Jun--Aug), and SON (Sep--Nov). Boundary-crossing windows remain whole; these are not exact calendar-season totals. Hours and episodes are observed; margin is standardized \$/home-week against fixed 20\%. Higher margin/promise is better. The released CSV retains all controllers/datasets and paired window-bootstrap intervals.}
\label{tab:seasonal}
\end{table}
\FloatBarrier
\endgroup

\section{Named storm windows and synthetic warnings}
\label{app:events}

The alert override uses county/zone hurricane and tropical-storm watches or warnings; severe-thunderstorm, tornado, winter-storm, blizzard, ice-storm, high-wind, squall, extreme-cold, wind-chill, and flash-flood warnings; and winter-weather advisories. The qualifying phenomena and significance codes are frozen in the released alert loader. The risk classifiers use separate, broader tropical, convective, winter, wind, and flood indicators.

Named event endpoints are exclusive. Uri spans February 13--20, 2021; Elliott December 22--26, 2022; derecho May 16--19, 2024; Beryl July 8--14, 2024; and August heat August 19--21, 2024, in America/Chicago. All event dates and the additional 2023 and 2025 stress windows are frozen in the public split file. Event rows retain their actual durations; their weekly margins are episode rates and should not be interpreted as forecast annual profits. A restored interval within a merged household episode can recharge the battery, as it does in the simulator.

\begin{figure}[htbp]
\centering
\includegraphics[width=\linewidth]{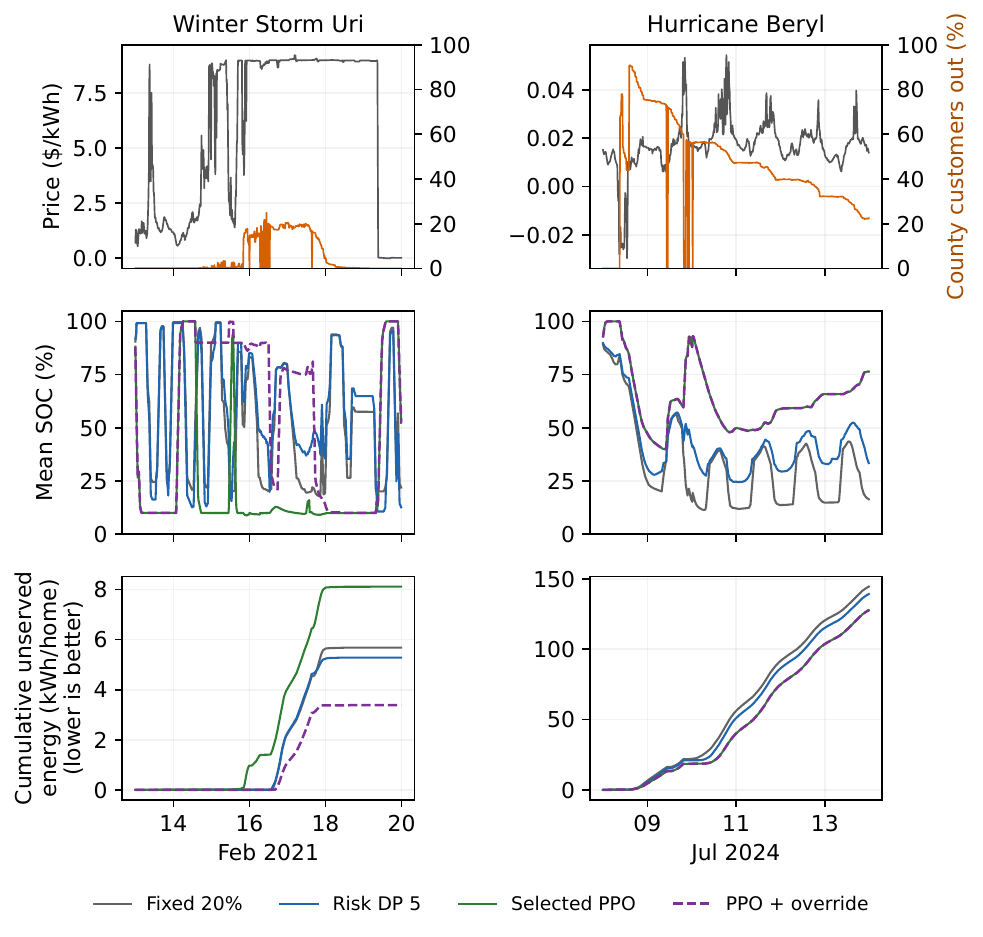}
\caption{Houston storm traces under identical sampled homes and outages. Top: observed price (gray, left axis) and recorded county outage fraction (orange, right axis). Middle: average fleet SOC, which can conceal individual depleted homes. Bottom: cumulative unserved energy per sampled home, lower is preferable. Full-outage unserved energy can continue rising after a five-hour commitment has already succeeded. Time is local.}
\label{fig:traces}
\end{figure}

\begin{table}[htbp]
\centering\small
\begin{tabular}{llrrrr}
\toprule
Window & Controller & Margin/wk & Promise (\%) & Unserved & Episodes \\
\midrule
HOU: Uri & Fixed 20\% & -865.1 & 90.9 & 5.7 & 198 \\
HOU: Uri & Storm rule & -783.8 & 100.0 & 2.1 & 198 \\
HOU: Uri & Risk DP 5 & -845.4 & 94.4 & 5.3 & 198 \\
HOU: Uri & Selected PPO & -908.6 & 69.7 & 8.1 & 198 \\
HOU: Uri & PPO + override & -785.4 & 98.0 & 3.4 & 198 \\
HOU: Elliott & Fixed 20\% & 81.1 & 100.0 & 0.0 & 1 \\
HOU: Elliott & Storm rule & 5.1 & 100.0 & 0.0 & 1 \\
HOU: Elliott & Risk DP 5 & 85.8 & 100.0 & 0.0 & 1 \\
HOU: Elliott & Selected PPO & 12.5 & 100.0 & 0.0 & 1 \\
HOU: Elliott & PPO + override & 22.2 & 100.0 & 0.0 & 1 \\
HOU: Derecho & Fixed 20\% & 11.0 & 87.6 & 5.6 & 89 \\
HOU: Derecho & Storm rule & 10.9 & 87.6 & 5.5 & 89 \\
HOU: Derecho & Risk DP 5 & 10.0 & 92.1 & 5.2 & 89 \\
HOU: Derecho & Selected PPO & 6.7 & 96.6 & 4.2 & 89 \\
HOU: Derecho & PPO + override & 6.7 & 96.6 & 4.2 & 89 \\
HOU: Beryl & Fixed 20\% & 7.8 & 81.6 & 144.5 & 261 \\
HOU: Beryl & Storm rule & 7.8 & 83.1 & 141.5 & 261 \\
HOU: Beryl & Risk DP 5 & 7.3 & 96.9 & 139.3 & 261 \\
HOU: Beryl & Selected PPO & 7.0 & 97.7 & 127.8 & 261 \\
HOU: Beryl & PPO + override & 7.0 & 97.7 & 127.8 & 261 \\
DFW: Elliott & Fixed 20\% & 78.7 & 100.0 & 0.0 & 3 \\
DFW: Elliott & Storm rule & 78.7 & 100.0 & 0.0 & 3 \\
DFW: Elliott & Risk DP 5 & 84.8 & 100.0 & 0.0 & 3 \\
DFW: Elliott & Selected PPO & -3.3 & 100.0 & 0.0 & 3 \\
DFW: Elliott & PPO + override & -3.3 & 100.0 & 0.0 & 3 \\
\bottomrule
\end{tabular}

\caption{Named-window results. Margin is terminal-adjusted \$/home-week, promise is a percentage, and unserved energy is kWh/home over the actual window. Episode counts refer to simulated household episodes and remain common across controllers. An unavailable promise means no scored episode. Higher margin/promise and lower unserved energy are preferable.}
\label{tab:events}
\end{table}

\begin{table}[htbp]
\centering\small
\begin{tabular}{lrrrr}
\toprule
Threshold\% & Events & Active\% & Past 24 h\% & Lead $\ge4$ h\% \\
\midrule
1 & 455 & 35.8 & 47.9 & 2.9 \\
5 & 41 & 85.4 & 97.6 & 19.5 \\
\bottomrule
\end{tabular}

\caption{County warning coverage, all six evaluated counties, 2019--2025. Threshold is the customers-out percentage defining an audit event. ``Active'' means at onset; ``past 24 h'' means any recent qualifying warning and does not require continuity. The last column requires continuous active lead of at least four hours, the worst preset's charging time from its physical minimum. These fractions are not household service-success probabilities.}
\label{tab:alerts}
\end{table}

The synthetic design crosses three prices (50, 1,000, 5,000 \$/MWh), three outage durations (2, 8, 24 h), four warning leads (0, 1, 3, 6 h), and two initial SOC values (10\%, 90\%), giving 72 cases. Six false/missed-warning cases and eight load/SOH/dropout cases give 86 indexed cases in total. One nominal setting repeats as a reference in the robustness block, leaving 85 distinct parameter settings. Each indexed case is evaluated for an always-discharge proposal and selected PPO, both guarded and unguarded, giving 344 comparisons. Thirty-two homes use 25 kWh presets, demand is 2 kW before an optional twofold uplift, and synthetic outages affect a county fraction of 0.5. SOH sensitivity is 0.85 versus one and dropout sensitivity zero versus 0.2. Price remains high through the specified joint-event window; the full episode lasts 48 hours. These controlled perturbations are stress scenarios, not samples from a calibrated event generator.
\FloatBarrier

\section{Risk-forecast diagnostics}
\label{app:hazard}

\begin{table}[htbp]
\centering\small
\begin{tabular}{lrrrrr}
\toprule
Year & Horizon (h) & ROC AUC & Avg. precision & Brier & Positive\% \\
\midrule
2021 & 1 & 0.952 & 0.744 & 0.0114 & 2.50 \\
2021 & 6 & 0.842 & 0.443 & 0.0307 & 4.27 \\
2021 & 24 & 0.728 & 0.357 & 0.0767 & 9.73 \\
2024 & 1 & 0.963 & 0.815 & 0.0083 & 2.49 \\
2024 & 6 & 0.876 & 0.599 & 0.0225 & 4.00 \\
2024 & 24 & 0.774 & 0.415 & 0.0648 & 8.75 \\
2025 & 1 & 0.865 & 0.399 & 0.0072 & 1.02 \\
2025 & 6 & 0.758 & 0.228 & 0.0227 & 2.68 \\
2025 & 24 & 0.678 & 0.204 & 0.0702 & 8.06 \\
\bottomrule
\end{tabular}

\caption{Out-of-training-period county classifier diagnostics pooled across the twelve weather-covered Texas counties used to fit/evaluate the forecast. Larger ROC AUC/average precision and smaller Brier score are preferable. Positive prevalence differs by year/horizon. Evaluation fleets use six of these counties. Calibration on 2020 does not imply calibration under later distribution changes.}
\label{tab:hazard}
\end{table}

Each horizon classifier uses 300 histogram gradient-boosting iterations, learning rate 0.05, at most 31 leaves, balanced class weights, and seed zero. Isotonic calibration is fitted separately at each horizon. Calendar-year membership selects feature timestamps; future-window training labels can extend by at most 24 hours into the next calendar year. This boundary convention is shared by the saved forecast build and remains well before Uri for 2020 calibration. The bundle release freezes the resulting forecast arrays and fitted model.
\FloatBarrier

\section{Fleet sensitivities and runtime}
\label{app:structure}

\begin{table}[htbp]
\centering\small
\begin{tabular}{llrrr}
\toprule
Scenario & Dispatch & $\Delta$ margin/wk & Promise\% & Episodes \\
\midrule
base 100 & fixed20 & +0.0 & 85.9 & 608 \\
base 100 & storm & +7.2 & 89.8 & 608 \\
base 100 & risk5 & +3.3 & 94.9 & 608 \\
size 500 & fixed20 & +0.0 & 84.5 & 2966 \\
size 500 & storm & +11.5 & 88.7 & 2966 \\
size 500 & risk5 & +4.1 & 94.5 & 2966 \\
size 1000 & fixed20 & +0.0 & 84.4 & 5925 \\
size 1000 & storm & +10.4 & 88.5 & 5925 \\
size 1000 & risk5 & +4.5 & 94.0 & 5925 \\
core 70pct & fixed20 & +0.0 & 88.7 & 608 \\
core 70pct & storm & +6.4 & 92.1 & 608 \\
core 70pct & risk5 & +2.1 & 96.5 & 608 \\
dropout 20pct & fixed20 & +0.0 & 86.2 & 608 \\
dropout 20pct & storm & +14.9 & 89.8 & 608 \\
dropout 20pct & risk5 & +1.5 & 94.1 & 608 \\
base 100 & ppo & -14.3 & 88.0 & 608 \\
base 100 & ppo guard & +10.7 & 97.2 & 608 \\
size 500 & ppo & -5.8 & 87.8 & 2966 \\
size 500 & ppo guard & +13.5 & 97.3 & 2966 \\
size 1000 & ppo & -12.4 & 87.6 & 5925 \\
size 1000 & ppo guard & +12.9 & 97.5 & 5925 \\
core 70pct & ppo & -20.6 & 89.3 & 608 \\
core 70pct & ppo guard & +10.1 & 99.0 & 608 \\
dropout 20pct & ppo & -1.9 & 87.8 & 608 \\
dropout 20pct & ppo guard & +18.5 & 97.2 & 608 \\
\bottomrule
\end{tabular}

\caption{Six-window fleet sensitivity collection, not annual evaluation. Margin difference is \$/home-week against fixed 20\% under the same scenario and is weighted by duration. Promise pools household episode/failure counts across the six windows. Scenario names specify fleet size, a 70\% Core mix, or 20\% independent per-interval dropout; other settings match the base 100-home scenario. All sampled source pools contain 300 profiles.}
\label{tab:structure}
\end{table}

\begin{table}[htbp]
\centering\small
\begin{tabular}{rlrr}
\toprule
Homes & Dispatch & Mean ms/interval & SD \\
\midrule
100 & fixed20 & 1.80 & 0.02 \\
100 & risk5 & 2.18 & 0.11 \\
100 & ppo & 0.34 & 0.00 \\
100 & ppo guard & 0.35 & 0.02 \\
500 & fixed20 & 8.47 & 0.09 \\
500 & risk5 & 9.79 & 0.14 \\
500 & ppo & 0.40 & 0.01 \\
500 & ppo guard & 0.41 & 0.00 \\
1000 & fixed20 & 16.84 & 0.22 \\
1000 & risk5 & 19.84 & 0.11 \\
1000 & ppo & 0.49 & 0.00 \\
1000 & ppo guard & 0.51 & 0.02 \\
\bottomrule
\end{tabular}

\caption{Mean CPU wall time per interval for complete forecast/decision/simulation loops, excluding episode reset. Three measured one-day repetitions follow a warm-up for each configuration. The CSV records platform, backend, thread count and all repetitions. DP replans hourly with 32 grid points; PPO acts every interval. These are implementation timings, not communication delays or isolated neural-network inference times.}
\label{tab:latency}
\end{table}
\FloatBarrier

\section{Full static feeder audit}
\label{app:ac}

The feeder has 1,581 load elements. A fixed seed selects battery locations at each penetration; profiles cycle over the same 100 homes across controllers. Nonattached locations use the load/PV reference. We solve 96 quarter-hours after one day of battery replay on April 15 (normal), July 8 (Beryl), and August 20 (heat), 2024. Three controls at two penetrations plus three no-battery references yield 21 rows. Original voltage-dependent load models are retained. The voltage mask uses the source-bus graph of enabled lines and transformers; it does not certify conductor continuity or phase energization. Relay, protection, restoration, feeder feedback, and AC dynamics are omitted.

\begin{table}[htbp]
\centering\small
\setlength{\tabcolsep}{4pt}
\begin{tabular}{llrrrrr}
\toprule
Day & Dispatch & Pen.\% & Voltage range & UV\% & OV\% & Line\% \\
\midrule
beryl & Load/PV reference & 0 & [0.812, 1.108] & 5.2 & 9.4 & 3.1 \\
beryl & Fixed 20\% & 30 & [0.773, 1.108] & 14.6 & 22.9 & 24.0 \\
beryl & Risk DP 5 & 30 & [0.812, 1.108] & 15.6 & 19.8 & 25.0 \\
beryl & PPO + override & 30 & [0.812, 1.108] & 12.5 & 21.9 & 28.1 \\
beryl & Fixed 20\% & 100 & [0.793, 1.108] & 22.9 & 32.3 & 28.1 \\
beryl & Risk DP 5 & 100 & [0.813, 1.109] & 24.0 & 35.4 & 29.2 \\
beryl & PPO + override & 100 & [0.814, 1.110] & 19.8 & 30.2 & 36.5 \\
heat & Load/PV reference & 0 & [0.851, 1.084] & 17.7 & 17.7 & 39.6 \\
heat & Fixed 20\% & 30 & [0.791, 1.120] & 26.0 & 27.1 & 57.3 \\
heat & Risk DP 5 & 30 & [0.792, 1.120] & 30.2 & 34.4 & 61.5 \\
heat & PPO + override & 30 & [0.728, 1.134] & 33.3 & 49.0 & 56.2 \\
heat & Fixed 20\% & 100 & [0.000, 1.140] & 95.8 & 27.1 & 62.5 \\
heat & Risk DP 5 & 100 & [0.000, 1.141] & 91.7 & 28.1 & 63.5 \\
heat & PPO + override & 100 & [0.000, 1.137] & 70.8 & 36.5 & 56.2 \\
normal & Load/PV reference & 0 & [0.889, 1.084] & 5.2 & 5.2 & 4.2 \\
normal & Fixed 20\% & 30 & [0.888, 1.084] & 7.3 & 18.8 & 6.2 \\
normal & Risk DP 5 & 30 & [0.847, 1.084] & 9.4 & 18.8 & 8.3 \\
normal & PPO + override & 30 & [0.889, 1.131] & 5.2 & 21.9 & 20.8 \\
normal & Fixed 20\% & 100 & [0.888, 1.084] & 11.5 & 21.9 & 8.3 \\
normal & Risk DP 5 & 100 & [0.847, 1.084] & 11.5 & 29.2 & 11.5 \\
normal & PPO + override & 100 & [0.889, 1.193] & 5.2 & 21.9 & 20.8 \\
\bottomrule
\end{tabular}

\caption{Full AC audit. Penetration is the percent of load locations assigned battery dispatch. Voltage range is across phase nodes at graph-reachable buses and all intervals, using the mask above. UV/OV are percentages of intervals with any retained phase voltage below 0.95/above 1.05 pu; Line is the percentage with any line current above its nominal ampacity. A small minimum is retained even when the snapshot solver reports convergence. Eighteen phase nodes at unreachable buses have zero voltage and are counted separately in the released CSV. The reference's existing violations prevent treating controller changes as a calibrated grid-compliance assessment.}
\label{tab:ac}
\end{table}
\FloatBarrier

\section{Aging sensitivity conventions}
\label{app:aging}

Calendar fade is fixed at 2\% per year, independent of dispatch. The throughput proxy is $f_{\rm ET}=2.71\times10^{-5}$ times charge-plus-discharge energy divided by twice rated capacity. The cycle proxy is $\sum_j f_{\rm CB}d_j^{2.03}n_j$, with $f_{\rm CB}=5.24\times10^{-4}$, normalized SOC range $d_j$, and rainflow full/half-cycle count $n_j$. These coefficients follow \citet{humiston2026degradation} without recalibration to the commercial packs. Throughput and rainflow fades are annualized by observed duration.

BLAST-Lite 1.1.1 uses \texttt{Lfp\_Gr\_SonyMurata3Ah\_Battery}. Each SOC and completed-hour ambient-temperature trace is tiled and truncated to 365 days, avoiding linear scaling of short-window calendar fade while introducing repetition and boundary discontinuities. Ambient temperature proxies cell temperature; thermal management, chemistry differences, aging feedback, and stochastic long-term cycling are unmodeled. These columns describe model sensitivities under fixed dispatch, rather than calibrated pack-lifetime forecasts.

\begin{table}[htbp]
\centering\small
\begin{tabular}{lrrrr}
\toprule
Dispatch & Mean SOC (\%) & ET & Rainflow & BLAST \\
\midrule
fixed20 & 56.6 & 0.54 & 3.05 & 4.40 \\
hold90 & 80.5 & 0.39 & 2.13 & 5.75 \\
storm & 58.3 & 0.55 & 3.10 & 4.49 \\
risk5 & 65.3 & 0.50 & 2.51 & 4.93 \\
ppo & 77.8 & 0.38 & 2.17 & 5.40 \\
ppo guard & 77.8 & 0.38 & 2.18 & 5.40 \\
\bottomrule
\end{tabular}

\caption{Mean across four 20-home window sensitivities (Beryl, derecho, August heat, and regular window w10). ET and rainflow columns are annualized capacity-loss percentages; BLAST repeats each trace to one year. Equal weighting across these windows is a reporting convention, not a representative annual climate mix. Mean SOC is a percentage. Lower predicted fade is preferable within a model; values across model classes are not interchangeable.}
\label{tab:aging}
\end{table}
\FloatBarrier

\end{document}